\documentclass[10pt,twocolumn,letterpaper]{article}

\usepackage[table]{xcolor}
\usepackage{iccv}

\makeatletter
\@namedef{ver@everyshi.sty}{}
\makeatother
\usepackage{tikz}

\usepackage{amsmath, amsfonts, times, epsfig, amssymb,  blindtext, float, subcaption, multicol, array, multirow, graphicx, etoolbox, enumitem, accents, tabularx, tabu, comment, soul}
\usepackage{color}
\usepackage[makeroom]{cancel}

\usepackage{algorithmicx}
\usepackage[ruled, vlined, linesnumbered]{algorithm2e}
\SetKwProg{Fn}{Function}{}{}
\usepackage{mathtools}
\usepackage{setspace}

\usepackage{tabstackengine}
\setstackEOL{\cr}
\newcommand{\sota}{\emph{state-of-the-art}}
\newcommand{\sftb}{\emph{so-far-the-best}}

\newcommand{\rulesep}{\unskip\ \vrule\ }
\definecolor{headergray}{gray}{.9}

\usepackage{lipsum}
\newcommand\blfootnote[1]{%
  \begingroup
  \renewcommand\thefootnote{}\footnote{#1}%
  \addtocounter{footnote}{-1}%
  \endgroup
}

\usepackage[pagebackref=true,breaklinks=true,letterpaper=true,colorlinks,bookmarks=false]{hyperref}

\iccvfinalcopy 

\def\iccvPaperID{****} 

\begin{document}

\title{VSAC: Efficient and Accurate Estimator for H and F}


\author{Maksym Ivashechkin$^{1}$, Daniel Barath$^{2}$, and Jiri Matas$^{1}$\\
$^1$ Centre for Machine Perception, Czech Technical University in Prague, Czech Republic\\
$^2$ Computer Vision and Geometry Group, Department of Computer Science, ETH Zürich\\
{\tt\small \{ivashmak, matas\}@cmp.felk.cvut.cz} \:\:\: {\tt\small dbarath@ethz.ch}\\
}

\maketitle
\ificcvfinal\thispagestyle{empty}\fi



\begin{abstract}
We present VSAC, a RANSAC-type robust estimator with a number of novelties.
It benefits from the introduction of the concept of independent inliers that improves significantly the efficacy of the dominant plane handling and, also, allows near error-free rejection of incorrect models,  without false positives.
The local optimization process and its application is improved so that it is run on average only once.
Further technical improvements include  adaptive sequential hypothesis verification and  efficient model estimation via Gaussian elimination.
Experiments on four standard datasets show that 
VSAC is significantly faster than all its predecessors and
runs on average in 1-2 ms, on a CPU. It is two orders of magnitude faster and yet as precise as MAGSAC++, the currently most accurate estimator of two-view geometry.  In the repeated runs on EVD, HPatches, PhotoTourism, and Kusvod2 datasets, it never failed. 
\end{abstract}

\section{Introduction}
The {Ran}dom {Sa}mple {C}onsensus (RANSAC) 
algorithm introduced by Fischler and Bolles \cite{fischler1981random} is one of the most popular robust estimators in computer science.
The method is widely used in computer vision,
its applications include stereo matching \cite{torr1993outlier, torr1998robust}, image mosaicing \cite{ghosh2016survey}, motion segmentation \cite{torr1993outlier}, 3D reconstruction, detection of geometric primitives, and  structure and motion estimation~\cite{conf/iccv/Nister03}.

The textbook version of RANSAC proceeds as follows: random samples of minimal size sufficient to estimate the model parameters are drawn repeatedly.
Model consistency with input data is evaluated, \eg, by counting the points closer than a manually set inlier-outlier threshold. If the current model is better then the \sftb, it gets stored. 
The procedure terminates when the probability of finding a better model falls below a user-defined level. 
Finally, the estimate is polished by least-squares fitting of inliers.

Many modifications of the original algorithm have been proposed.
Regarding sampling,
PROSAC~\cite{chum2005matching} exploits an a priori predicted inlier probability rank.
NAPSAC \cite{nasuto2002napsac} samples in the neighborhood of the first, randomly selected, point. 
Progressive NAPSAC~\cite{barath2019progressive} combines both and adds gradual convergence to uniform spatial sampling.

In textbook RANSAC, the model quality is measured by its support, \ie, the number of inliers, points consistent with the model.
MLESAC~\cite{torr2000mlesac} introduced a quality measure that makes it the maximum likelihood procedure.
To avoid the need for a user-defined noise level, MINPRAN~\cite{stewart1995minpran} and A-contrario RANSAC~\cite{a-contrario} select the inlier-outlier threshold so that the inliers are the least likely to occur at random. 
Reflecting the inherent uncertainty of the threshold estimate, 
MAGSAC~\cite{barath2019magsac} marginalizes the quality function over a range of noise levels.
MAGSAC++~\cite{Barath_2020_CVPR} proposes an iterative re-weighted least-squares optimization of the {\sftb} model with weights calculated from the inlier probability of points.
The Locally Optimized RANSAC \cite{chum2003locally} refines the {\sftb}  model using a non-minimal number of points, \eg, by iterated least-squares fitting. 
Graph-Cut RANSAC \cite{barath2018graph}, in its local optimization, exploits the fact that real-world data tend to form spatial structures.
The model evaluation is usually the most time-consuming part as it depends both on the number of models generated and the number of input data points. 
A quasi-optimal speed-up was achieved by the Sequential Probability Ratio Test (SPRT) \cite{matas2005randomized} that randomizes the verification process itself.

In many cases, points in degenerate configuration affect the estimation severely. For example, correspondences lying on a single plane is a degenerate case for $\mathbf{F}$ estimation. 
DEGENSAC \cite{chum2005two} detects such cases and applies the plane-and-parallax algorithm. 
USAC \cite{raguram2013usac} was the first framework integrating many of the mentioned techniques, including PROSAC, SPRT, DEGENSAC, and LO-RANSAC. 

In this paper,  we present VSAC%
\footnote{VSAC has multiple novelties and we found no natural abbreviation reflecting them. We chose "V" as the letter following "U", as in USAC.
}%
, a RANSAC-type estimator that exploits a number of novelties. It is significantly faster than all its predecessors, and yet as precise as MAGSAC++, the currently most accurate method both in our experiments and according to a recent survey~\cite{ma2021image}. The accuracy reaches, or is very near, the geometric error of the ground truth, estimated by cross-validation.
For homography $\mathbf{H}$ and epipolar geometry $\mathbf{F}$ estimation,
VSAC runs on average in 1-2 ms (on a CPU) on all datasets,
two orders of magnitude faster than MAGSAC++. In the repeated runs on datasets EVD~\cite{mishkin2015mods}, HPatches~\cite{Balntas_2017_CVPR}, PhotoTourism~\cite{phototourism} and Kusvod2~\cite{lebeda2012fixing}, it never failed. 

Moreover, VSAC is able to reject non-matching image pairs, with a zero false positive rate on hundreds of random image pairs and a zero false negative rate on pairs from the above-mentioned datasets. 
The ability is underpinned by a novel concept of {\it independent random inliers} in the contrario context.
We show that if dependent random inliers, \eg spatially co-located points, are not counted, the support of random models follows very closely a Poisson distribution with a single parameter $\lambda$ that is easy to estimate reliably%
\footnote{For geometric problems, the Poisson distribution is a tight approximation of the binomial. Moreover, only the mean $\lambda$ of independent random inlier counts needs to be estimated, instead of $T$ (number of trails) and $\delta$ (success probability) for the binomial. 
} 
for the given pair. 
The easily calculated CDF of Poisson raised to the power of the number of evaluated models provides the probability that a certain model quality was reached by chance.
VSAC thus provides two confidence measures together with its result. The first is the classical one -- the probability that RANSAC returned the model with the highest support. 
The second is the confidence that the returned solution was not obtained by chance.

The concept of independent random inliers plays a critical role in VSAC's accuracy and robustness. 
Experiments show that most failures of USAC-like methods for $\mathbf{F}$ estimation occur in the presence of a dominant plane, despite the DEGENSAC algorithm.
In such cases, there are only few out-of-plane inliers, and due to structures in the outliers, incorrect models with high support exist. Removing the dependent structures addresses the problem. Further improvements of dominant plane handling include a heuristic guess of the calibration matrix allowing to deal with fully planar scenes and detects pure rotation. If the guess is wrong, the support reveals it and nothing but a microsecond is lost.

The speed of VSAC is achieved with several technical improvements. Most significantly, we attack the problem of expensive local optimization. In the LO-RANSAC paper~\cite{chum2003locally}, the authors prove that the local optimization is run at most $\log(k)$ times, where $k$ is the number of iteration. Nevertheless, despite $ \log(k) \ll k$, the complex local optimization may end up being the efficiency bottleneck. We show that a fast local optimization combined with {\it a single complex final optimization} leads to a faster, yet equally precise algorithm. Moreover, by detecting the intersection over union of {\sftb} and the current set of inliers and by not optimizing similar models, an algorithm is obtained that runs the local optimization on average about once and almost always fewer than two times. 

Further speed up is gained by adaptive SPRT. We use the estimated expected number of random inliers to tune SPRT~\cite{matas2005randomized} to the outlier density of the processed pair. We also measure, on the fly, the actual time of model estimation and model verification on the given hardware at the given moment, which is needed for calculating the quasi-optimal thresholds of the SPRT.

To broaden its application potential, VSAC provides novel outputs. Employing the highly efficient Lindstrom method for triangulation \cite{triangulationEasyLindstrom}, it obtains the point pair exactly fitting the returned $\mathbf{F}$ that minimizes the geometric error. VSAC can be thus employed for noise filtering. VSAC returns all input points sorted by the residual, allowing the user to set his own ex-post trade-off between the density, and possibly spread, of points on the one hand, and their accuracy on the other. 

\section{Detecting Random Models}
\label{sec:randomness}

\begin{figure}[t]
    \centering
    \includegraphics[width=0.45\columnwidth]{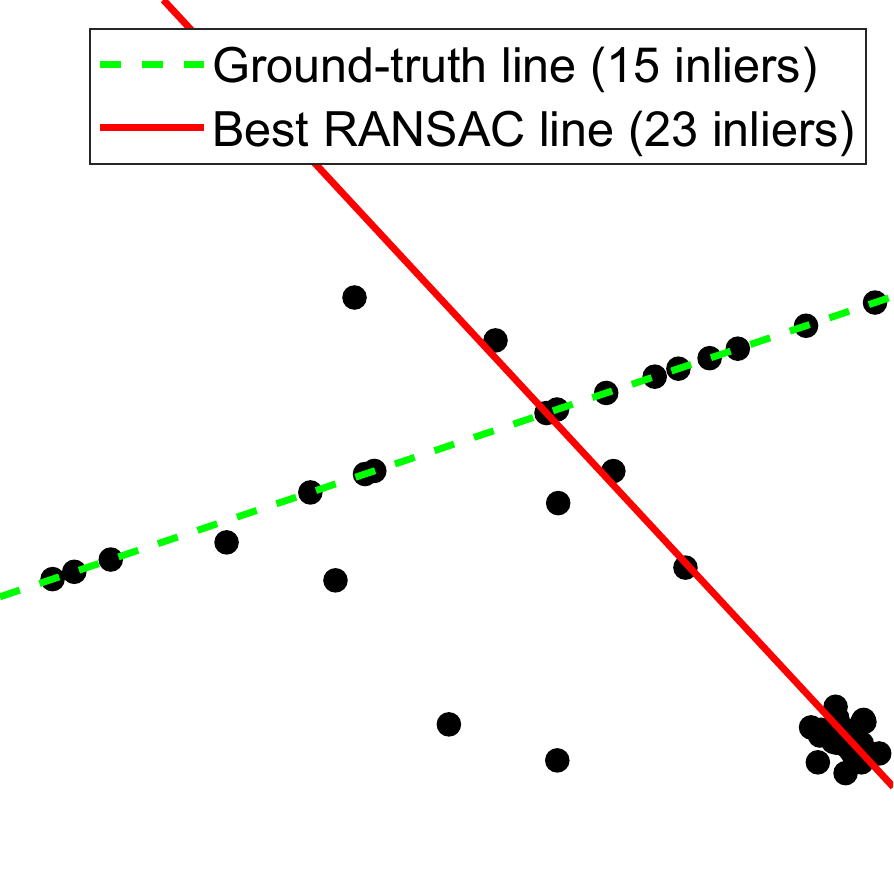} \hspace{1mm} \rulesep \hspace{1mm}
    \includegraphics[width=0.45\columnwidth]{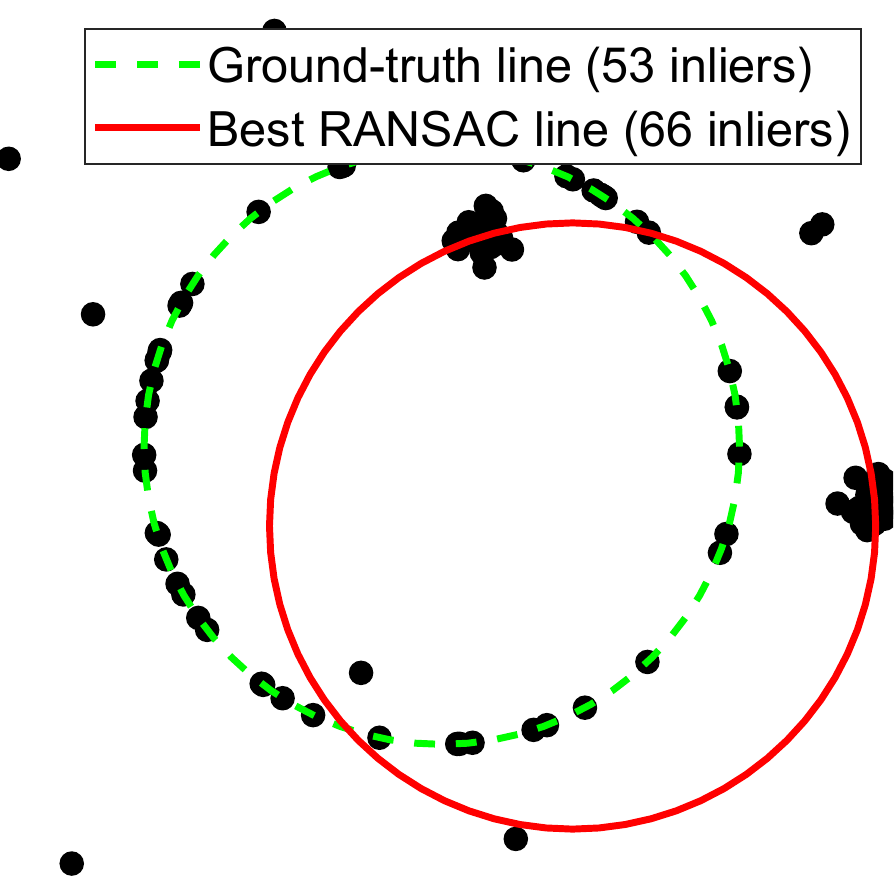}
    \vspace{-0.8em}
    \caption{Examples of random models found by RANSAC with a large number of dependent inliers. }
    \label{fig:example_random_model}
\end{figure}

One issue of RANSAC-like robust estimators is the inability to recognize failures.  
The estimator always returns a model maximizing some quality function, \eg the inlier count, even if the tentative inliers stem from outlier structures -- sets of neighboring data points that do not originate from the sought model manifold and significantly affect the quality function when considered inliers. In such cases, the returned model might have a reasonably large number of inliers while being inconsistent with the underlying scene geometry. See Fig.~\ref{fig:example_random_model} for examples.

In this section, we propose a new approach for detecting failures.
Towards that end, we differentiate between \textit{independent} and  \textit{dependent} inliers. This split is conceptual, helping exposition -- 
in the a contrario calculation of the probability,  we pick one point in a group of structured points and count it as an inlier arising by chance, it is an independent random inlier.
The other inliers in the structure are ignored, since their inlier status is not a random event, but rather a consequence of their spatial dependence and the fact that the independent inlier is consistent. A \emph{non-random} model must have a sufficient number of \emph{independent} inliers.

We define data point ${\bf p}$ a \textit{dependent} inlier if its point-to-model residual is smaller than the inlier-outlier threshold and one of the following conditions hold.
\begin{enumerate}[noitemsep]
    \item Point ${\bf p}$ is in the minimal sample used for estimating the model parameters. In such cases, the point will have zero residual by definition. 
    \item Point ${\bf p}$ is close to an independent inlier ${\bf q}$, $||{\bf p} - {\bf q}|| \rightarrow 0$.
    In such cases, points ${\bf p}$ and ${\bf q}$ form a spatial structure that affects the model quality significantly. 
    Thus, only point ${\bf q}$ is considered independent random. 
    Other points from the structure, \eg, { \bf p}, are dependent inliers.
\end{enumerate}
These conditions are valid for general data points and model to be estimated. 
In case of estimating epipolar geometry from point correspondences, we define the following additional conditions as well. 
\begin{enumerate}[noitemsep]
    \setcounter{enumi}{2}
    \item A correspondence $\left({\bf p}, {\bf p}^{\prime} \right)$ where ${\bf p}$ or ${\bf p}^{\prime}$ is close to the epipole in the corresponding image is considered dependent since $\left({\bf p}, {\bf p}^{\prime} \right)$ always satisfies epipolar constraint ${\bf p}^{\prime\top} \mathbf{F} {\bf p} = 0$. 
    This stems from the fact that $\mathbf{F} {\bf p} \backsimeq \mathbf{F} {\bf e} = {\bf 0}$ if $||{\bf p} - {\bf e}|| \rightarrow 0$, where ${\bf e}$ is the epipole in the first image. The same holds in the second one.
    \item Correspondence $\left({\bf p}, {\bf p}^{\prime} \right)$ is a dependent inlier if it does not pass the chirality check~\cite{chum2004epipolar}. 
    \item Let $(\bf{l}, \bf{l}^\prime)$ be the corresponding epipolar lines of an independent inlier correspondence.
    All correspondences that are closer to lines $(\bf{l}, \bf{l}^\prime)$ than the inlier-outlier threshold are considered dependent.
\end{enumerate}
Data points that have a point-to-model residual smaller than the inlier-outlier threshold and do not satisfy any of the previous conditions are considered independent inliers.

To decide if a model estimated by RANSAC is random and inconsistent with the underlying scene geometry and, thus, should be considered failure,
we use the univariate theory~\cite{univariate-Sandilands2014}. 
Suppose that we are given $N$ models estimated inside RANSAC during its run and the corresponding numbers of random independent inliers $I_1, I_2, \dots, I_N \in \mathbb{N}$.
The number of $I$ points being consistent with a random model follows the binomial distribution.
The sequence of inliers is an \textit{i.i.d.}\ random variable with cumulative binomial distribution $C_{\mathcal{B}}(T, \delta)$, where $T$ is the number of points and $\delta$ is the probability that a point is an independent inlier to a bad (random) model.
The distribution of $I_{\max} = \max\{I_1, I_2, \dots, I_N\}$ over $N$ models is $C_{\mathcal{B}}(T, \delta)^N$. 
In order to recognize a good (non-random model) with confidence $p \rightarrow 1$, the following condition must hold.
\begin{equation}
  C_{\mathcal{B}}(I_{\max}; T, \delta)^N \ge p.
\label{eqn:success}
\end{equation}
In our experiments, we found that probability $\delta$ is fairly low.
Therefore, even a few independent inliers are enough to decide whether a model is random or not.
In this case, the binomial distribution can be approximated by Poisson distribution which is faster to compute. 
The only parameter of the Poisson distribution is $\lambda = T \, \delta$ that represents the mean number of independent inliers.

Finding independent inliers is relatively expensive and, therefore, it is computationally prohibitive if done for every model generated inside RANSAC.
Hence, we estimate parameter $\lambda$ from the first $n \ll N$ generated models. 
Parameter $\lambda$ is the mean number of independent inliers consistent with a bad model. 
In RANSAC, all models are considered bad that have fewer inliers than the {\sftb} one. 
However, good models that are slightly worse than {\sftb} can corrupt the estimation of $\lambda$.
Parameter $\lambda$ is found as follows:
first, we discard the {\sftb} model and the ones that have significantly overlapping inlier sets with it (measured via Jaccard similarity~\cite{jaccard}) from the generated $n$ models.
Second, $\tilde{\lambda}$ is robustly found via taking the median of \emph{independent} inlier counts of $n$ models.
Afterwards, we compute 95\% percentile ($I^{\%}_{95}$) of the Poisson distribution with parameter $\Tilde{\lambda}$ to filter out models with high supports. 
Eventually, $\hat{\lambda}$ is estimated as the average of inlier numbers lower than $I^{\%}_{95}$.

Finally, to decide if the model returned by RANSAC is non-random and, thus, should be accepted, the cumulative
distribution of  $(\max\{I_1, \dots, I_N\}; \hat{\lambda})^N$ is calculated, and the model is considered a good one if \eqref{eqn:success} holds.

\section{DEGENSAC$^+$}
\label{degensac_section}

Chum \etal~\cite{chum2005two} proposed the DEGENSAC algorithm 
that handles estimation, in an uncalibrated setup, 
of epipolar geometry  in scenes with a dominant plane.
Such scenes often appear in real-world scenarios, especially, in a man-made environment containing, \eg, building facades. 
In brief, after a minimal sample is selected to estimate the model parameters, DEGENSAC first checks whether correspondences in the sample are co-planar.
If at least five of them are, the corresponding $\mathbf{H}$ is estimated and, used in $\mathbf{F}$ calculation by the plane-and-parallax algorithm~\cite{hartley2003multiple}.

When experimenting with DEGENSAC, we found the following issues that we address in this section.
{First}, we observed that having five or six correspondences consistent with a homography does not necessarily mean that the estimated fundamental matrix is degenerate. 
This stems from the fact that the inlier-outlier threshold, used for selecting co-planar correspondences via thresholding the re-projection error, is usually wide enough to consider slightly curved surfaces to be planar.
{Second}, the plane-and-parallax algorithm applied for recovering $\mathbf{F}$ from a homography and two additional correspondences may fail, \eg, when the scene is entirely planar and when a large number of random inliers supports an incorrect $\mathbf{F}$. 

To solve these issues, we propose the DEGENSAC$^+$ algorithm by incorporating the inlier randomness criteria in the procedure. 
Both of the above-mentioned problems are solved by checking if the estimated epipolar geometry has a reasonable number of independent out-of-plane inliers. 
This builds on the observation that if a fundamental matrix is degenerate due to a dominant planar structure, it only has out-of-plane inliers by chance -- otherwise, it would not be degenerate. 
Thus, it is enough to check the number of independent inliers that are inconsistent with the homography. 

\textbf{Calibrated DEGENSAC$^+$.} 
The plane-and-parallax algorithm that is used in DEGENSAC to recover from a degenerate sample assumes that a reasonable number of out-of-plane correspondences exists and the camera motion is not rotation-only. 
In case one of these conditions does not hold, the algorithm fails while often requiring many RANSAC iterations to recognize this failure.

To solve this issue, we propose to use the intrinsic calibration matrices $\mathbf{K}_1, \mathbf{K}_2 \in \mathbb{R}^{3 \times 3}$ normalizing the coordinates of the co-planar correspondences in the minimal sample before estimating homography $\mathbf{H}$.
In such cases, $\mathbf{H}$ can be decomposed to rotation $\mathbf{R} \in \text{SO(3)}$ and translation $\mathbf{t} \in \mathbb{R}^3$~\cite{homogrDecomp} that can be then used to find the sought non-degenerate fundamental matrix.
When decomposing $\mathbf{H}$, we are given two candidate rotations $\mathbf{R}_a$ and $\mathbf{R}_b$ and translations $\pm {\bf t}_a, \pm {\bf t}_b$.
Since the relative pose can only be estimated up-to-scale, $-{\bf t}_a$ and $-{\bf t}_b$ can be rejected, and the candidate solutions are $(\mathbf{R}_a, {\bf t}_a)$ and $(\mathbf{R}_b, {\bf t}_b)$. The solution is selected with the maximum inlier count.
The final fundamental matrix is
%
$\mathbf{F} = \mathbf{K}^{-\top}_2 [\hat{{\bf t}}]_{\times} \hat{\mathbf{R}} \mathbf{K}^{-1}_1$,
%
where $\hat{\mathbf{R}} \in \{ \mathbf{R}_a, \mathbf{R}_b \}$ and $\hat{{\bf t}} \in \{ \mathbf{t}_a, \mathbf{t}_b \}$.
With this solution, the rotation-only cases are straightforwardly detected. 

In many cases, the calibration is not known a priori but can be approximated.
Setting the principal points to coincide with the image centers is a widely used approach. 
Estimating the focal length is, however, a more challenging problem. 
We found that testing a range of candidate focal lengths to find a reasonable approximation leads to a reliable $\mathbf{F}$ estimate while still being faster than the original plane-and-parallax RANSAC applied inside DEGENSAC.

The proposed algorithm incorporating the non-randomness criteria proposed in the previous section and also the calibration matrix is shown in Alg.~\ref{alg:degensac}. 
\begin{algorithm}[t]
\setstretch{0.92}
\SetArgSty{textnormal}
\DontPrintSemicolon
  \KwIn{$\hat{\mathbf{F}}^*$ -- {\sftb} fundamental matrix; $\mathcal{S}$ -- minimal sample initializing $\hat{\mathbf{F}}^*$, $I_F$ -- initial guess for independent non-planar support. }
        $\hat{\mathbf{H}}^*$ := \emph{estimateHomography} ($\hat{\mathbf{F}}^*, \mathcal{S}$)\;
        \uIf{$\hat{\mathbf{H}}^* = \varnothing$}{
            {\bf return: } $\hat{\mathbf{F}}^*$ \CommentSty{// no homography} \;
        }
        \uIf{$\mathbf{K}$ \emph{is given}}{
            \emph{assert} ($\mathbf{K}^{-1}\hat{\mathbf{H}}^*\mathbf{K}$ is not conj. to rotation) \;
            $\hat{\mathbf{F}}$ := \emph{findF} ($\hat{\mathbf{H}}^*$, $\mathbf{K}$) \;
        } \Else {
            $\hat{\mathbf{F}}^\prime$ := \emph{findF} ($\hat{\mathbf{H}}^*$, $\hat{\mathbf{K}}$) \CommentSty{// use approx.$\,\hat{\mathbf{K}}$} \;
        }
    \uIf{\emph{support} ($\hat{\mathbf{F}}^*) \ge I_F$}
    {
          {\bf return: } $\hat{\mathbf{F}}^*$ \CommentSty{// not degenerate} \;
    }
    \uIf{$\mathbf{K}$ \emph{is given}}{
        {\bf return: } $\hat{\mathbf{F}}$  \CommentSty{// cannot be degenerate}  \;
    }
    \uIf{\emph{support}$(\hat{\mathbf{F}}^\prime) \ge I_F$}
    {
          {\bf return: } $\hat{\mathbf{F}}^\prime$\;
    } 
    \CommentSty{// re-estimate support $I_F$} \;
    $\hat{\mathbf{F}}^{\prime\prime}, I_F$ := \emph{planeAndParallax} ($\hat{\mathbf{H}}^{*}$) \;
    \uIf{\emph{support} $(\hat{\mathbf{F}}^{\prime}) \ge I_F$ \text{or} \emph{support} $(\hat{\mathbf{F}}^{\prime\prime}) \ge I_F$}
    {
          {\bf return: } $\hat{\mathbf{F}}^{\prime}$ or $\hat{\mathbf{F}}^{\prime\prime}$\;
    } 
    {\bf return: } $\varnothing$ \CommentSty{// reject $\hat{\mathbf{F}}^{*}$}
\caption{DEGENSAC$^+$}
\label{alg:degensac}
\end{algorithm}

\section{Adaptive SPRT}
The Sequential Probability Ratio Test (SPRT) proposed by Matas \etal \cite{matas2005randomized} aims at speeding up the robust estimation procedure by addressing the problem that, in RANSAC, a large number of models are verified, \eg their support is calculated, even if they are unlikely to be better than the previous {\sftb}. 
The time spent on these models is wasted. 
The SPRT test is based on Wald's theory of sequential decision making. 
It interrupts the model verification when the probability of that particular model being a \emph{good} one falls below a user-defined threshold. 

SPRT has four user-defined parameters, \ie, 
the initial probability of a correspondence being consistent with a \emph{good} ($\epsilon_0$) and a \emph{bad} model ($\delta_0$); 
avg.\ number of estimated models ($\bar{m}_{\mathcal{S}}$); time to estimate the model parameters ($t_M$). 
The actual parameters that lead to the fastest procedure are challenging to find manually and require the user to acquire knowledge about the problem at hand. 
Even the architecture of the computer impacts the minimal solver and point verification times that should be considered when setting $t_M$.
To avoid the manual setting, we propose the Adaptive SPRT (A-SPRT) algorithm that finds the optimal SPRT parameters in a data- and architecture-dependent manner.

\noindent
\textbf{The model estimation time} depends not only on the computer architecture but, also, on the actual solver and error metric being used. Parameter $t_M$ is calculated for free by measuring the model estimation and point verification run-times in the first $n$ RANSAC iterations. 

\noindent
\textbf{The average number} of models $\bar{m}_{\mathcal{S}}$ is found as the average number of \emph{valid} models per sample in the first $n$ iterations. 

\noindent
\textbf{Inlier probabilities} $\epsilon_0$ and $\delta_0$ are estimated from the 
average number $\hat{\lambda}$ of inliers consistent with a bad model 
that is estimated from in the first $n$ RANSAC iterations as $\delta_0 = \hat{\lambda} \,/\, T$, see Section~\ref{sec:randomness}. 
Parameter $\delta_0$ is the probability of a point being an inlier and $\hat{\lambda}$ is the mean of the corresponding binomial distribution $\mathcal{B}(T, \delta_0)$.
From $\delta_0$, we approximate the maximum number of  inliers $I_{\delta}$ of a \emph{bad} model as a high quantile (\eg, 0.99) of the normal distribution with the same mean and standard deviation as $\mathcal{B}$. The approximation is as follows:
\begin{equation}
    I_{\delta} = \hat{\lambda} + 3.719\,\sqrt{\hat{\lambda}\,(1-\delta_0})
    \label{eqn:min_inliers}
\end{equation}
The initial probability $\epsilon_0$ of a correspondence being inlier of a \emph{good} model is found as $\epsilon_0 = \max(I_{\delta}, \hat{I}^{*}) \:/ \:T$,
%
%
where $\hat{I}^{*}$ is the inlier number of the {\sftb} model.

\noindent
\textbf{Failure case.} 
If probabilities $\delta_0$ and $\epsilon_0$ are similar, the original SPRT often rejects \emph{good} models leading to increased run-time or, in extreme situations, total failure.
To solve this issue, we propose to apply A-SPRT only if
\begin{equation}
    \frac{1}{1 - \alpha}\,t^{w}_v\,E^{w}(T) < t_{v}\,T,
\end{equation}
where $t^{w}_v$ and $t_v$ are the times for verifying a single correspondence, respectively, with and without SPRT and $\alpha$ is the probability of a false rejection~\cite{matas2005randomized}, 
and $E^{w}(T)$ is the average number of points verified. 

\section{Model Accuracy}
\label{lo_section}
\noindent
\textbf{Simple and Fast Local Optimization.}
The key idea of the local optimization (LO) proposed in~\cite{chum2003locally} is to address the fact that not all all-inlier samples lead to accurate models due to the noise in the data.
We however observed that the LO step has a slightly different role in practice -- finding a model that is good enough to trigger the termination criterion of RANSAC early.
The final model accuracy depends mostly on the optimization procedure, \eg least-squares fitting or numerical optimization, applied once, after the RANSAC main loop finished.  
Therefore, the LO can be made light-weight without compromising the final model accuracy.

In our formulation, the primary objective is to find a light-weight LO procedure that runs swiftly and is applied only when it likely leads to termination. 
To do so, we introduce the following conditions that control when the LO is applied. 
They are as follows: 
\begin{enumerate}[noitemsep]
    \item 
    \textit{Non-random model}. The {\sftb} model has the required number of independent inliers defined by \eqref{eqn:min_inliers}.
    \item \textit{Low Jaccard similarity.} 
    The local optimization is applied only if the inlier sets of the new best and previous best models has a lower than $0.95$ Jaccard index, \ie, the intersection over union.
    This condition is motivated by the tendency that if a model is just slightly different from the previous {\sftb}, the LO step likely does not refine it significantly, but the final optimization does the main improvement.
\end{enumerate}

\begin{algorithm}[t]
\DontPrintSemicolon
  \KwIn{$\mathcal{I}^*$ -- inliers of {\sftb} model, \par$\tau^*$ -- score of {\sftb} model, \texttt{MAX\_ITERS} -- maximum iterations of LO,\: $s$ -- sample size. }

   $\theta^{*}_{\text{LO}} := \varnothing$; \: $I_{\max} := |\mathcal{I}^*|$ \;
    \For{$t := 0;\:\: t < \emph{\texttt{MAX\_ITERS}}; \:\:t$++}
    {
        $\theta_{\text{LO}} := $ \emph{estimate (subset($\mathcal{I}^*$, $\min$\{$|\mathcal{I}^{*}|, s$\}))} \;
        $\tau_{\text{LO}}, \mathcal{I}_{\text{LO}} :=$ \emph{evaluate($\theta_{\text{LO}}$)}\;
        \uIf{$\tau^{*} \prec \tau_{\text{LO}}$ \& \emph{non-degenerate($\theta_{\text{LO}}$)}}
        {
             $\mathcal{I}^{*}, \tau^*, \theta^{*}_{\text{LO}} := \mathcal{I}_{\text{LO}}, \tau_{\text{LO}}, \theta_{\text{LO}}$\;
        }
        \uIf{$I_{\max} < |\mathcal{I}_{\text{LO}}|$}
        {
            $I_{\max} := |\mathcal{I}_{\text{LO}}|$ \;
        }
    }
{\bf return:} $\theta^*_{\text{LO}}$, $\tau^*$, \emph{MaximumIterations($I_{\max}$)}\;
\caption{Local Optimization}
\label{alg:LO}
\end{algorithm}

The proposed LO procedure is shown in Alg.~\ref{alg:LO}. 
It is important to note that in this procedure, larger-than-minimal samples are selected that is typically avoided in RANSAC due to increasing the problem complexity and, thus, the number of iterations required to provide probabilistic guarantees of finding the sought model parameters.
In Alg.~\ref{alg:LO}, the sample is selected from a set of points that likely are inliers.
Therefore, the increased sample size does not affect the accuracy and processing time negatively. 

We found experimentally the parameters that suit for two-view geometric problems, minimizing the total run-time while maintaining the accuracy.
For $\mathbf{F}$ estimation the optimal sample size is 21 and the number of iterations is 20.
For $\mathbf{H}$, the optimal size is 32 and number of iterations is 10. 

\noindent
\textbf{Final Optimization.} 
Since the proposed local optimization does not intend to make the {\sftb} model as accurate as possible, the final model polishing step should return the best possible model. 
In textbook RANSAC~\cite{fischler1981random}, the final model parameters are obtained by running a single least-squares fitting on the set of inliers returned by RANSAC.
 
In our experiments, we test two types of final optimizations. The first is an iterative LSQ fitting that re-selects the inliers.
This improves the model parameters extremely efficiently.
The second one is the iteratively re-weighted LSQ approach proposed in~\cite{Barath_2020_CVPR} that does not require a single inlier-outlier threshold, only its loose upper bound.
This is, in practice, slightly slower than traditional iterative LSQ. 
Due to applying it only once, the deterioration in the run-time is at most 0.2-1.0 milliseconds in our experiments. 

\noindent
\textbf{Minimal Model Estimation} 
often requires finding the null-space of a linear system, typically, by SVD.
The traditionally used solvers for $\mathbf{H}$/$\mathbf{F}$/$\mathbf{E}$ (homography, fundamental, essential) matrix estimation use null space parameterization either to directly solve the problem ($\mathbf{H}$ from 4 point correspondences) or to find the coefficients of some polynomials that are then solved ($\mathbf{F}$ from 7 matches; $\mathbf{E}$ from 5 matches)~\cite{hartley2003multiple}.
It is however slow when having large matrices. Given that it is applied in every RANSAC iteration, it affects the total run-time severely. 
Instead of SVD, we suggest to use Gaussian Elimination (GE) as in~\cite{RHO}. 
While GE is less stable numerically in theory, it has many advantages, \eg, an order-of-magnitude speed-up, easy-to-implement, and lower memory complexity than SVD.
The marginally decreased stability of the minimal solver does not affect the final accuracy.
\section{Point Correction and Ranking} 
RANSAC outputs the model with the highest support and the corresponding set of inliers.
Some applications, \eg, 3D reconstruction or bundle adjustment, rely heavily on the obtained set of inliers, and it is thus important to rank them according to the quality.
A simple way is to sort the inliers by their residuals in an increasing order.

Correcting the inliers by making them consistent with the estimated model is an extremely important problem, \eg, for improving data points provided as the ground truth. For instance, such points are obtained by careful manual selection in a number of real-world datasets, \eg \textsc{Kusvod2} and \textsc{AdelaideRMF}~\cite{wongiccv2011}, that leads to correct, however, inevitably noisy points.
Even if we assume that the annotator did a perfect work and all found points are supposedly noise-free, the discrete nature of photography (\ie, the scene is projected to a grid of pixels) prevents having perfect points with no noise. 
Also, correcting points is very useful for the user who does not want to run bundle adjustment.
We therefore propose to minimize the noise by correcting a point so it has a zero error distance (\eg, reprojection distance for $\mathbf{H}$ or geometric error for $\mathbf{F}$) to the final model.

\noindent
\textbf{Homography.}
A way to correct correspondences is by introducing ``half'' homography $\mathbf{A}$, where $\mathbf{H} = \mathbf{A}\mathbf{A}$ \cite{square_root}.
The middle point averaging out the points in this reference frame is calculated as follows:
%
    ${\bf m} = (\phi(\mathbf{A}{\bf x}) + \phi(\mathbf{A}^{-1}{\bf x}^{\prime}))/ 2$,
%
where $\phi$ is a mapping, normalizing a point by its homogeneous coordinate.
The points projected to the manifold are $\tilde{{\bf x}} \sim \mathbf{A}^{-1}{\bf m}$ and $\tilde{{\bf x}}^{\prime} \sim \mathbf{A}{\bf m}$.
%
%
The up-to-scale relation $\sim$ can be removed via mapping $\phi$. The corrected correspondence $(\tilde{{\bf x}}, \tilde{{\bf x}}^{\prime})$ has zero error to $\mathbf{H}$, because the elimination of ${\bf m}$ from the two equations implies $\tilde{{\bf x}}^\prime \sim \mathbf{AA}\tilde{{\bf x}} = \mathbf{H} \tilde{{\bf x}}$. The square root $\mathbf{A}$ exists if the eigenvalues of $\mathbf{H}$ have positive real parts. Planar homographies from image datasets satisfy this condition in the experiments. In general, a homography should be checked before applying the proposed point correction.

\noindent
\textbf{Epipolar geometry.}
The corrected points must lie \emph{perfectly} on the epipolar lines.
A fast procedure was presented by Lindstrom in \cite{triangulation_lindstrom}. 
Moreover, if the intrinsic camera matrices are known, \cite{triangulation_lindstrom} enables to efficiently obtain the triangulated 3D points as well.
Since \cite{triangulation_lindstrom} corrects the correspondences even they are incorrect matches, we use the oriented epipolar constraint \cite{chum2004epipolar} to remove some of the incorrect ones. 
Results are put in the appendix.

\begin{table}[tb]
\setlength{\tabcolsep}{4.6pt}
\renewcommand{\arraystretch}{.9}
\begin{center}
\begin{tabular}{c|c||r|r|r|r}
\hline \rowcolor{headergray}
 &  \multirow{2}{*}{} &  \multicolumn{4}{c}{\# of RANSAC iterations} \\ \cline{3-6}\rowcolor{headergray} 
\multirow{-2}{*}{Problem}& & $\sim 10^2$ & $\sim 10^3$ & $\sim 10^4$ & $\sim 10^5$ \\ \hline\hline
\multirow{2}{*}{$\mathbf{H}$} & w & \textbf{100\%} & \textbf{100\%} & \textbf{100\%} & \textbf{100\%} \\  \cline{2-6} 
                 & w/o & 96\% & 93\% & 86\% & 81\%
\\ \hline\hline
\multirow{2}{*}{$\mathbf{F}$} & w & \textbf{100\%} & \textbf{99\%} & \textbf{99\%} & \textbf{100\%} \\ \cline{2-6}
   & w/o & 85\% & 82\% & 79\% & 78\% \\ \hline
\end{tabular}
\end{center}
\vspace{-1.5em}
\caption{Percentage of detected failures by the proposed criterion \eqref{eqn:success} with (w) and without (w/o) removing dependent inliers on homography $\mathbf{H}$ and fundamental matrix $\mathbf{F}$ estimation problems when trying to match image pairs without a common field-of-view, \ie, they do not match. In total, 500 image pairs tested.
}
\label{fig:failures}
\end{table}
\begin{table}[tb]
\begin{center}
\renewcommand{\arraystretch}{0.825}
\setlength{\tabcolsep}{3.0pt}
\begin{tabular}{l|c|c|c|c}
\hline \rowcolor{headergray} 
 \multicolumn{1}{c|}{Dataset}                     & Method & \multicolumn{1}{c|}{Error (px)} & \multicolumn{1}{c|}{$t_{\text{total}}$ (ms)} & \multicolumn{1}{c}{$t_{\text{DEG}}$ (ms)} \\ \hline
\multirow{2}{*}{PhotoTourism} & DEG\phantom{$^+$} & 0.64 & 54.7 & 19.8 \\ \cline{2-5} 
& DEG$^+$ & \textbf{0.44} & \textbf{31.6} & \phantom{1}\textbf{1.7}\\ \hline
\multirow{2}{*}{Kusvod2}     & DEG\phantom{$^+$} & 2.34 & 19.4 & \phantom{1}5.7 \\ \cline{2-5} 
& DEG$^+$ & \textbf{1.83} & \textbf{11.6} & \phantom{1}\textbf{1.4} \\ \hline
\end{tabular}
\vspace{-0.6em}
\caption{
Avg.\ geometric error (px), RANSAC run-time ($t_{\text{total}}$; ms), and time for recovering from degenerate solutions ($t_{\text{DEG}}$; ms) of DEGENSAC$^+$ and DEGENSAC~\cite{chum2005two} on image pairs from datasets \textsc{Kusvod2} (7 pairs) and PhotoTourism~\cite{lebeda2012fixing,phototourism} (100) where $1/3$ of the correspondences are consistent with a homography, \ie, dominant plane.
 }
\label{table:degensac}
\end{center}
\end{table}

\section{Experiments}
To test VSAC and each of the new techniques proposed in this paper, we have downloaded the \textsc{EVD} (11 pairs; avg.\ inlier ratio: $0.24 \pm 0.16$) \cite{mishkin2015mods} and \textsc{HPatches} (142; ratio: $0.51 \pm 0.24$) \cite{Balntas_2017_CVPR} datasets for $\mathbf{H}$, and the
\textsc{Kusvod2} (7; ratio: $0.52 \pm 0.18$) \cite{lebeda2012fixing} and 
\textsc{PhotoTourism} (500 pairs are chosen randomly; ratio: $0.37 \pm 0.12$) \cite{phototourism} for $\mathbf{F}$ estimation. 

\textbf{DEGENSAC$^+$} is compared to DEGENSAC on images from the \textsc{Kusvod2} and \textsc{PhotoTourism} datasets where at least $1/3$ of the correspondences are consistent with a homography, \ie, dominant plane. 
For \textsc{PhotoTourism}, we used the ground truth camera calibration.
For \textsc{Kusvod2}, the intrinsic parameters are approximated as proposed in Section~\ref{degensac_section}
The avg.\ error (px), overall RANSAC time, and the run-times of the DEGENSAC versions (ms) are reported in Table~\ref{table:degensac}. 
DEGENSAC$^+$ leads to the lowest errors while being faster than the original DEGENSAC algorithm.


\textbf{Local Optimization} criteria proposed in Section~\ref{lo_section} are tested on the four downloaded datasets. 
Table~\ref{tab:loruns} reports the average number of samples drawn, {\sftb} models encountered and LO runs. 
While the original LO-RANSAC~\cite{chum2003locally} applies LO always when a new {\sftb} model is found, the proposed criteria leads to running LO significantly less often, \ie, once per problem, causing a significant speed-up with no deterioration in the accuracy. 

\begin{figure}[t]
    \centering
    \includegraphics[width=0.48\linewidth]{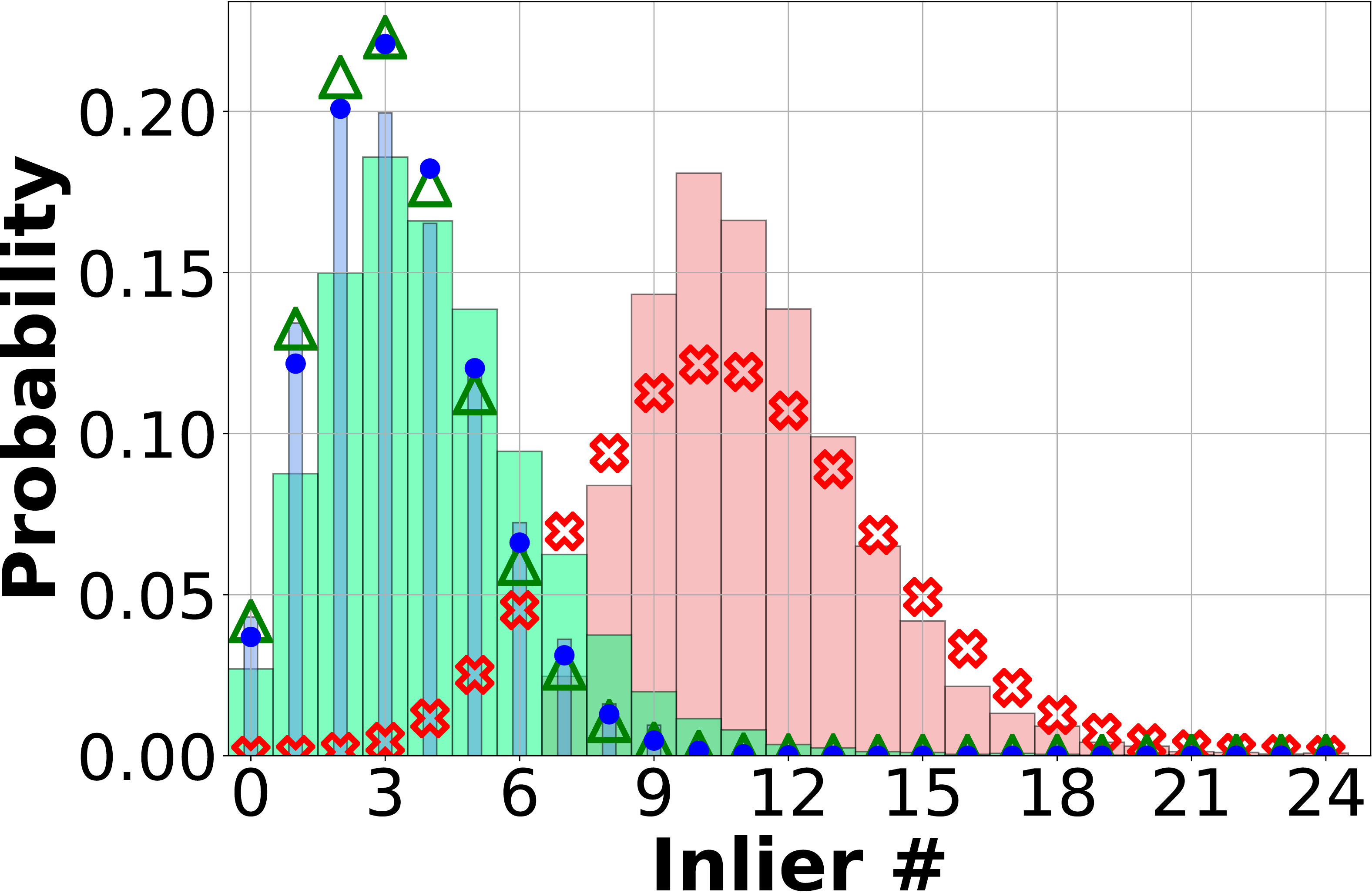}
    \includegraphics[width=0.47\linewidth]{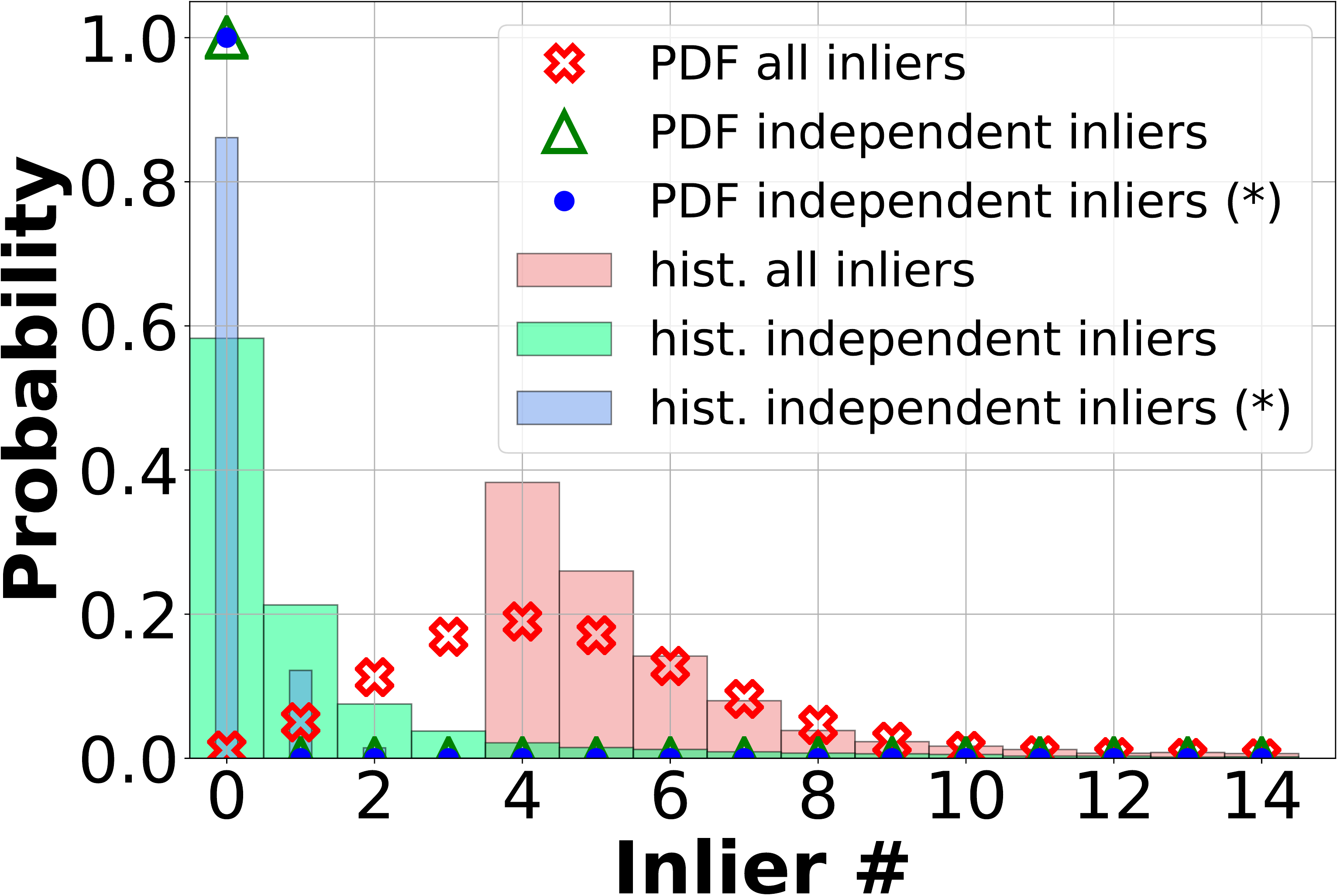}
    \vspace{-0.5em}
    \caption{
    Histograms of the numbers of all (red bar) and independent (green bar) inliers of models estimated in RANSAC; and the expected inlier numbers from the Poisson distribution with its parameters calculated from all (red crosses) and from the independent inliers (green triangle) on a scene from \textsc{EVD} (left) an \textsc{PhotoTourism} (right) datasets.
    The blue histogram (*) reports inlier counts of models generated from corrupted samples. Thus, the blue PDF shows the true probability of points being consistent with a bad model.
    }
    \label{fig:poisson_match_pairs}
\end{figure}

\textbf{Dependent inlier} removal helps in modelling the probability of a model being random as shown in Fig.~\ref{fig:poisson_match_pairs}.
We estimated the parameters of the Poisson distribution (see Section \ref{sec:randomness}) on two scenes from the \textsc{PhotoTourism} and \textsc{EVD} datasets from all inliers (red crosses) and, also, from only the independent ones (green triangles).
We also show the histogram of the actual inlier numbers of models estimated inside RANSAC.
The green histogram is closer to the values of the corresponding Poisson distribution than the red one.
Thus, removing dependent inliers helps in recognizing and rejecting random models.


\textbf{Failure detection} results are shown in Table~\ref{fig:failures}. 
Images with no common scene were matched with SIFT~\cite{Lowe04distinctiveimage} detector having, thus, only incorrect correspondences, and therefore, no good model exists.
We calculated the percentage of cases when the proposed criterion \eqref{eqn:success} detects if the currently matched image pair has no common field-of-view, \ie, they do not match, with and without removing the dependent inliers.  
Criterion \eqref{eqn:success} with removing the dependent inliers almost \textit{always} detects if two images do not match.


\begin{table}[tb]
\renewcommand{\arraystretch}{.9}
\centering
\begin{tabular}{l|r|r|r}
\hline  \rowcolor{headergray} 
\multicolumn{1}{c|}{Dataset} & \multicolumn{1}{c|}{Sample $\#$} & \multicolumn{1}{c|}{Sftb $\#$} & \multicolumn{1}{c}{LO $\#$} \\ \hline
HPatches    & 77.6 & 4.41 & 1.00 \\ \hline
EVD         & 534.1 &  5.66 & 1.20 \\ \hline
PhotoTourism & 14.4 & 3.49 & 1.00 \\ \hline
Kusvod      & 442.7 & 4.07 & 1.08 \\ \hline
\end{tabular}
\vspace{-0.6em}
\caption{The avg.\ number of samples drawn, {\sftb} (sftb) models encountered and LO runs on four datasets by VSAC.
For the original LO-RANSAC, the numbers of sftb and LO runs are identical.
VSAC runs notably fewer LO. 
}
\label{tab:loruns}
\end{table}

\definecolor{gold}{HTML}{BD820B}
\definecolor{silver}{HTML}{909090}
\definecolor{bronze}{HTML}{9A5F26}
\newcommand*\circledd[1]{\tikz[baseline=(char.base)]{
            \node[shape=circle,draw, inner sep=0.15pt] (char) {#1};}}           
\newcommand*\circleedd[1]{\tikz[baseline=(char.base)]{ \node[shape=box,fill=red!30, sep=0.0pt] (char) {#1};}}           
\newcommand*\circleeds[1]{\tikz[baseline=(char.base)]{ \node[shape=circle,draw,fill=white, inner sep=0.30pt](char){#1};}}           
\newcommand{\ffirst}[1]{%
    {\raisebox{0.8pt}{\circledd{\footnotesize\color{gold}{1}}} \hfill #1}%
}
\newcommand{\ssecond}[1]{%
    {\raisebox{0.8pt}{\circledd{\footnotesize\color{silver}{2}}} \hfill #1}%
}
\newcommand{\tthird}[1]{%
    {\raisebox{0.8pt}{\circledd{\footnotesize \color{bronze} {3}}} \hfill #1}%
}

\newcommand{\probclr}{\cellcolor{gray!5}}

\newcommand{\first}[1]{%
{\cellcolor{red!30}
{{\hspace{-1.5ex}\circleeds{\footnotesize{1}}}\hspace{3ex}\hfill{}#1}}}%
\newcommand{\second}[1]{%
{\cellcolor{cyan!20}
{{\hspace{-1.5ex}\circleeds{\footnotesize{2}}}\hspace{3ex}\hfill{}#1}}}%
\newcommand{\third}[1]{%
    {{\circleeds{\footnotesize {3}}} \hfill #1}%
}

\begin{table*}
\centering
\renewcommand{\arraystretch}{0.92}
\begin{tabular}{l|l|l|r|r|r|r||r|r|r|r}
\hline\rowcolor{headergray}
\multicolumn{1}{c}{\multirow{2}{*}{}}&
\multicolumn{1}{c|}{\multirow{2}{*}{}} &  & \multicolumn{4}{c||}{Time \hspace{3ex}(milliseconds)}   & \multicolumn{4}{c}{Error  \hspace{3ex}(pixels)} \\ \cline{4-11} \rowcolor{headergray}
\multicolumn{1}{c}{\multirow{2}{*}{}}&
\multicolumn{1}{c|}{} & \multicolumn{1}{c|}{\multirow{-2}{*}{Method}} & \multicolumn{1}{c|}{$t_{\text{med}}$} & \multicolumn{1}{c|}{$t_{\text{avg}}$} & 
\multicolumn{1}{c|}{$t_{\max}$} & \multicolumn{1}{c||}{$w^{t}_{\%}$} & \multicolumn{1}{c|}{$\varepsilon_{\text{med}}$} & \multicolumn{1}{c|}{$\varepsilon_{\text{avg}}$} &  \multicolumn{1}{c|}{$\varepsilon_{\max}$} & \multicolumn{1}{c}{$w^{\varepsilon}_{\%}$} \\ \hline\hline
\multirow{18}{*}{\rotatebox{90}{\textbf{Homography}}} &
\multirow{9}{*}{\rotatebox{90}{HPatches \hspace{2ex}(142 pairs)}}
&\textbf{VSAC} & \first{\phantom{00}0.8} & \first{\phantom{00}0.9} & \first{\phantom{00}3.0} & \first{\phantom{0}97} & 0.50 & 0.79 & 6.48 & 6\\ \cline{3-11}
&&\textbf{VSAC$_\text{MGSC}$}  & 1.8 & \second{\phantom{00}1.9} & \second{\phantom{00}6.9} & 0 & \first{0.43} & \first{0.61} & \first{3.47} & 16\\ \cline{3-11}
&&USACv20  & 2.8 & 3.3 & 9.1 & 0 & 0.51 & 8.16 & 509.19 & 6\\ \cline{3-11}
&&USAC  & 22.7 & 26.7 & 92.9 & 0 & 0.62 & 13.24 & 509.19 & 8\\ \cline{3-11}
&&OpenCV  & \second{\phantom{00}1.6} & 7.7 & 43.8 & \second{\phantom{00}3} & 0.58 & 1.99 & 141.99 & 9\\ \cline{3-11}
&&GC  & 174.3 & 157.8 & 337.6 & 0 & \first{0.43} & \second{0.62} & \second{4.34} & \second{23}\\ \cline{3-11}
&&MGSC++  & 20.3 & 44.1 & 987.5 & 0 & \second{0.48} & 0.68 & 8.08 & \first{24}\\ \cline{3-11}
&&ORSA  & 436.2 & 794.6 & 4692.7 & 0 & 0.82 & 105.69 & 4226.99 & 8 \\ \cline{3-11}
&& \multicolumn{5}{l||}{\probclr Cross-validation error on the ground truth points:} & \probclr 0.63 & \probclr 0.63 & \probclr 1.15 & \probclr \\ \cline{2-11}
&\multirow{9}{*}{\rotatebox{90}{EVD \hspace{2ex}(11 pairs)}}
&\textbf{VSAC} & \first{\phantom{00}0.5} & \first{\phantom{00}0.7} & \first{\phantom{00}2.9} & \first{100} & 3.18 & 3.54 & 7.65 & \second{18}\\ \cline{3-11}
&&\textbf{VSAC$_\text{MGSC}$}  & \second{\phantom{00}0.7} & \second{\phantom{00}0.9} & \second{\phantom{00}3.2} & 0 & \first{2.73} & \first{3.40} & \second{7.59} & 9\\ \cline{3-11}
&&USACv20  & 2.3 & 4.4 & 12.5 & 0 & 3.37 & 3.67 & 8.04 & \first{25}\\ \cline{3-11}
&&USAC  & 7.9 & 11.4 & 29.6 & 0 & 7.23 & 121.44 & 474.01 & 3\\ \cline{3-11}
&&OpenCV  & 16.9 & 21.1 & 40.1 & 0 & 4.02 & 5.36 & 16.17 & 5\\ \cline{3-11}
&&GC  & 31.6 & 33.8 & 52.3 & 0 & \second{2.78} & \second{3.48} & 10.42 & \first{25}\\ \cline{3-11}
&&MGSC++  & 28.4 & 27.0 & 63.2 & 0 & 3.48 & 3.78 & \first{7.54} & 13\\ \cline{3-11}
&&ORSA  & 56.8 & 68.1 & 218.9 & 0 & 148.79 & 169.66 & 438.45 & 4 \\ \cline{3-11}
&& \multicolumn{5}{l||}{\probclr Cross-validation error on the ground truth points:} & \probclr 1.80 & \probclr 2.21 & \probclr 6.33 & \probclr  \\ \hline \hline 
\multirow{20}{*}{\rotatebox{90}{\textbf{Fundamental matrix}}} &
\multirow{10}{*}{\rotatebox{90}{PhotoTour \hspace{.7ex}(500 pairs)}}
&\textbf{VSAC} & \first{\phantom{00}2.4} & \first{\phantom{00}2.5} & \first{\phantom{00}6.0} & \first{\phantom{0}99} & \first{0.15} & \first{0.16} & \second{0.64} & 12\\ \cline{3-11}
&&\textbf{VSAC$_\text{MGSC}$}  & \second{\phantom{00}3.0} & \second{\phantom{00}3.2} & \second{\phantom{00}7.7} & 0 & \first{0.15} & \second{0.17} & 0.70 & 12\\ \cline{3-11}
&&USACv20  & 20.0 & 29.4 & 100.3 & 0 & 0.17 & 0.20 & 3.37 & 10\\ \cline{3-11}
&&USAC  & 6.3 & 6.7 & 17.5 & 0 & 0.43 & 0.60 & 9.91 & 2\\ \cline{3-11}
&&OpenCV  & 170.0 & 149.3 & 280.1 & 0 & 0.38 & 0.63 & 10.00 & 1\\ \cline{3-11}
&&GC  & 224.2 & 256.9 & 696.4 & 0 & 0.17 & 0.19 & 0.93 & 9\\ \cline{3-11}
&&MGSC++  & 216.1 & 273.8 & 10764.9 & 0 & \second{0.16} & \second{0.17} & 0.93 & 17\\ \cline{3-11}
&&ORSA  & 84.0 & 98.9 & 594.0 & 0 & \first{0.15} & \first{0.16} & 0.95 & \first{19}\\ \cline{3-11}
&&NG-RSC  & 120.6 & 121.9 & 459.2 & 0 & \first{0.15} & \second{0.17} & \first{0.55} & \second{18} \\ \cline{3-11}
&& \multicolumn{5}{l||}{\probclr Cross-validation error on the ground truth points:} & \probclr 0.06 & \probclr 0.06 & \probclr 0.16 & \probclr  \\ \cline{2-11}
&\multirow{10}{*}{\rotatebox{90}{Kusvod2 \hspace{2ex}(15 pairs)}}
&\textbf{VSAC} & \first{\phantom{00}1.4} & \first{\phantom{00}2.5} & \first{\phantom{0}12.7} & \first{\phantom{0}45} & 0.53 & \second{1.06} & \first{6.37} & 11\\ \cline{3-11}
&&\textbf{VSAC$_\text{MGSC}$}  & 1.8 & 3.0 & \second{\phantom{0}13.1} & 0 & 0.53 & \first{1.00} & \second{6.39} & 9\\ \cline{3-11}
&&USACv20  & 2.3 & 7.6 & 75.4 & 3 & 0.56 & 2.85 & 54.52 & 12\\ \cline{3-11}
&&USAC  & \second{\phantom{00}1.5} & \second{\phantom{00}2.6} & 48.3 & \first{\phantom{0}45} & 2.11 & 3.37 & 36.46 & 3\\ \cline{3-11}
&&OpenCV  & 9.0 & 44.3 & 198.0 & \second{\phantom{00}7} & 1.02 & 4.52 & 38.11 & 0\\ \cline{3-11}
&&GC  & 189.7 & 214.5 & 893.0 & 0 & 0.54 & 1.49 & 35.66 & \first{23}\\ \cline{3-11}
&&MGSC++  & 47.4 & 84.4 & 1551.9 & 0 & \second{0.50} & 2.54 & 47.92 & \second{17}\\ \cline{3-11}
&&ORSA  & 32.4 & 62.0 & 605.6 & 0 & 0.53 & 15.43 & 255.56 & 10\\ \cline{3-11}
&&NG-RSC  & 105.1 & 110.4 & 599.6 & 0 & \first{0.48} & 3.27 & 39.84 & \second{17}\\ \cline{3-11}
&& \multicolumn{5}{l||}{\probclr Cross-validation error on the ground truth points:} & \probclr 0.91 & \probclr 1.12 & \probclr 2.34 & \probclr  \\  \hline
\end{tabular}
\vspace{-0.10cm}
\caption{
Comparison of speed and accuracy of the RANSAC methods. Average $t_{\text{avg}}$, median $t_{\text{med}}$, and the maximum $t_{\text{max}}$ running time over all runs, including 10 repetitions for EVD, HPatches and Kusvod2, on four datasets, measured on a standard CPU, with the exception of the NG-RSC methods, that requires a GPU.
Top and the second best results are marked and highlighted.
Similarly, we report average $\varepsilon_{\text{avg}}$, median $\varepsilon_{\text{med}}$ and the maximum $\varepsilon_{\text{max}}$ error of ground truth points.
To account for the randomization, the percentage of top results is given in the $w^{t}_{\%}$, $w^{\varepsilon}_{\%}$ columns.
The errors of the top methods are very close to the accuracy of ground truth points, estimated using leave-one-out. For Kusvod2, the best methods are significantly more precise than the ground truth, making the small differences meaningless.
}
\label{table:final_res}
\end{table*}





\textbf{Adaptive SPRT} is compared to the original algorithm proposed in \cite{matas2005randomized} and, also, to textbook RANSAC without preemptive verification on the \textsc{HPatches} and \textsc{PhotoTourism} datasets.
We chose a total of 480 image pairs that have inlier ratio lower than $0.5$. 
This was done to test the techniques on cases where RANSAC is required to do many iterations and, thus, the preemptive model verification is essential.

\begin{figure}[t]
    \centering
    \includegraphics[width=0.45\linewidth]{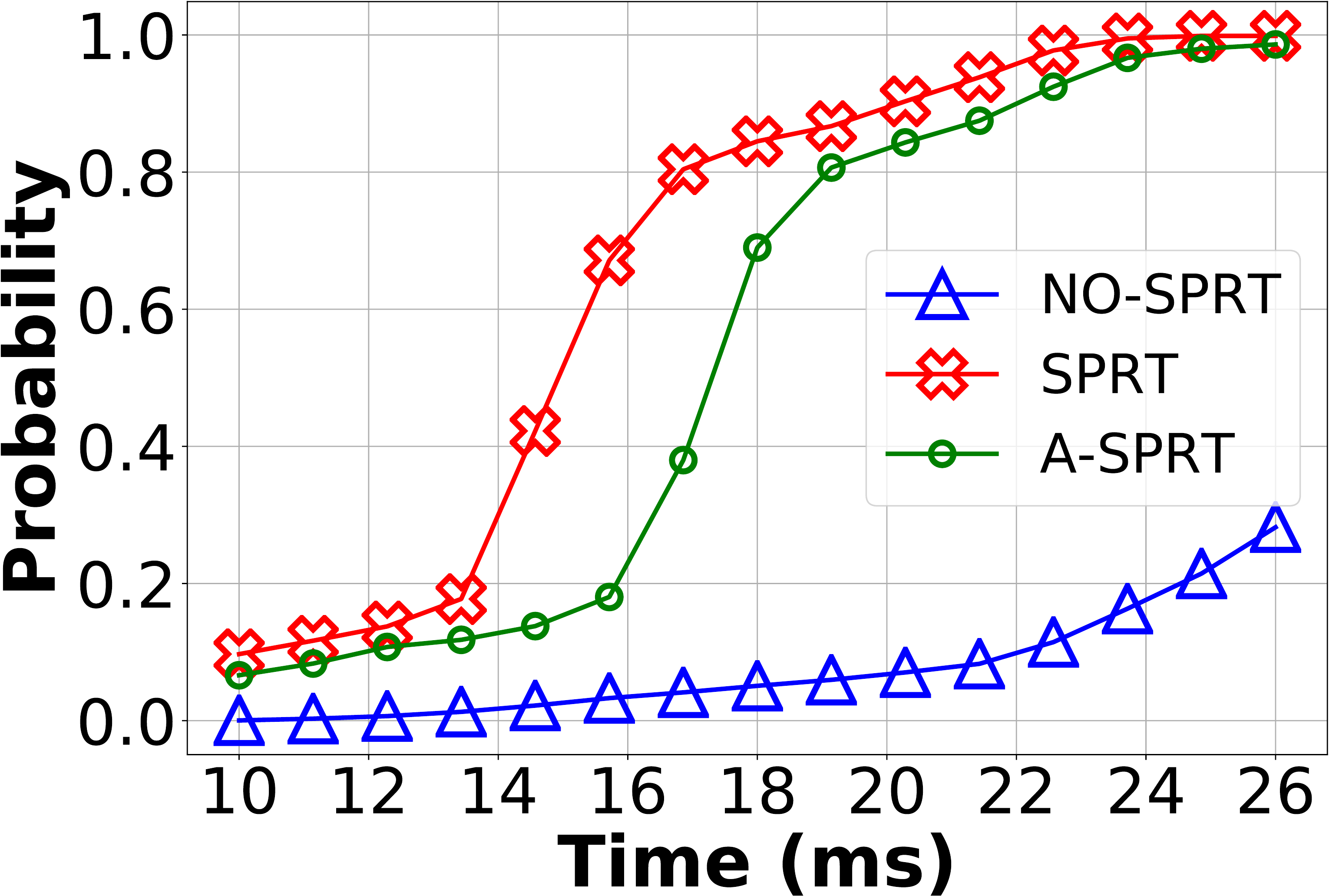}
    \includegraphics[width=0.45\linewidth]{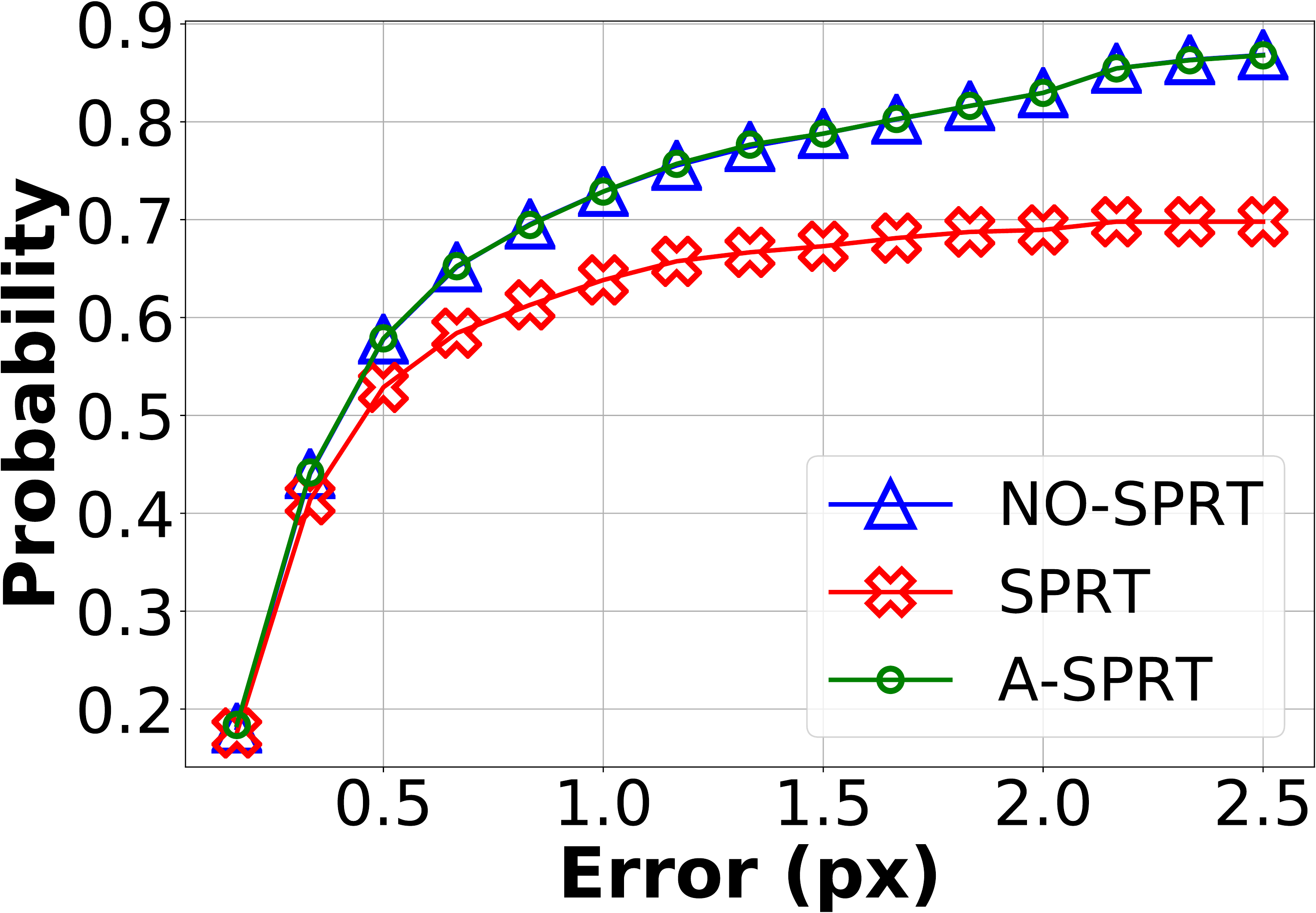}
    \vspace{-0.5em}
    \caption{
    CDFs of the processing times (ms) and geometric errors (px) of the proposed A-SPRT, SPRT, and RANSAC.}
    \label{fig:asprt}
\end{figure}

The CDFs of the estimation errors (in pixels) and processing times (in milliseconds) are shown in Fig.~\ref{fig:asprt}.
Being accurate is interpreted by a curve close to the top-left corner. 
It can be seen that, while the proposed A-SPRT leads to similar processing time as the original one (left plot), it is significantly more accurate (right).  

\textbf{Geometric Accuracy and Speed.} The proposed VSAC incorporating A-SPRT, Calibrated DEGENSAC$^+$, fast LO, PROSAC, and final covariance-based polishing (VSAC) or IRLS from \cite{Barath_2020_CVPR} (VSAC$_\text{MGSC}$) is compared with the following {\sota} robust estimators.
\textbf{(1)} USACv20 -- framework from \cite{USACv20} with SPRT, GC-RANSAC, DEGENSAC and P-NAPSAC. 
\textbf{(2)} default OpenCV's RANSAC implementation. 
\textbf{(3)} USAC framework from \cite{raguram2013usac} with SPRT, LO-RANSAC, DEGENSAC and PROSAC. 
\textbf{(4)} GC-RANSAC from \cite{barath2018graph} with SPRT, PROSAC and DEGENSAC. 
\textbf{(5)} MAGSAC++ from \cite{Barath_2020_CVPR} with P-NAPSAC and DEGENSAC.
\textbf{(6)} ORSA -- RANSAC with a contrario approach~\cite{a-contrario}.
\textbf{(7)} Neural-Guided RANSAC~\cite{brachmann2019ngransac} for epipolar geometry. 

The inlier-outlier threshold is set to 2.5 pixels for $\mathbf{H}$ and 1.5 pixels for $\mathbf{F}$; and the confidence to 99\%. The maximum number of iterations of the main loop of RANSAC is 3000 for homography and 5000 for fundamental matrix estimation. 
Table~\ref{table:final_res} shows that the proposed VSAC is \textit{always} the fastest methods with a large margin. Its maximum run-time ($t_{\text{max}}$) is 13 ms on a wide range of problems. 
The VSAC$_\text{MGSC}$ with the IRLS from \cite{Barath_2020_CVPR} as a final model optimization is slightly more accurate than VSAC while being marginally slower.
Both VSAC and VSAC$_\text{MGSC}$ have comparable accuracy to the {\sota} methods with many times being the most accurate methods.

For all methods, the source code provided by the authors were used. The methods are implemented in C++ except NG-RANSAC. The weight prediction part of NG-RANSAC runs in Python and CUDA on GPU. After the weight prediction, the rest of the code is in C++. All methods were run on the same computer, except NG-RANSAC. We ran it on a different machine with a GPU. 
\section{Conclusions}
This paper presented VSAC, a novel robust geometry estimator. We introduced the concept of independent inliers that helps detecting incorrect models,
and it is also at the core of the new DEGENSAC$^+$ method that returns a non-degenerate fundamental matrix supported by a sufficient number of non-planar independent inliers. 
The LO process was reformulated and sped-up while the novel final optimization is responsible for high geometric accuracy.
Technical enhancements as adaptive-SPRT with automatic parameter settings and GE for the minimal model solver provide further speed up.
VSAC is as geometrically precise as the best {\sota} methods.
\blfootnote{{\bf Acknowledgement.} This research was supported by the Research Center for Informatics (project CZ.02.1.01/0.0/0.0/16\_019/0000765 funded by OP VVV and by the Grant Agency of the Czech Technical University in Prague, grant No. SGS20/171/OHK3/3T/13.}


{\small
\bibliographystyle{ieee_fullname}
\bibliography{egbib}
}
\clearpage
\definecolor{headergray}{gray}{.9}
\definecolor{rowgray}{gray}{.95}
\newlength{\Oldarrayrulewidth}
\newcommand{\Cline}[2]{%
  \noalign{\global\setlength{\Oldarrayrulewidth}{\arrayrulewidth}}%
  \noalign{\global\setlength{\arrayrulewidth}{#1}}\cline{#2}%
  \noalign{\global\setlength{\arrayrulewidth}{\Oldarrayrulewidth}}}

\newcommand{\cclr}{\cellcolor{gray!10}}
\newcommand{\best}{\cellcolor{red!30}}
\newcommand{\snd}{\cellcolor{cyan!20}}
\section*{Appendix}
\section*{Point Correction}

We demonstrate how correcting the ground truth point correspondences, as proposed in Section 6, affects the results of the tested methods.
To do so, we corrected the ground truth correspondences provided in datasets EVD and HPatches (homography estimation), and in Kusvod2 and PhotoTour (fundamental matrix estimation). 
The results of the methods for homography and fundamental matrix estimation are shown in Table \ref{table:corrected_gt_errors}. 

In all cases, using the ground truth corrected by being projected to the model manifold, reduces the median and average errors of the tested method, allowing more accurate comparison. For $\varepsilon_{\max}$, the error is dominated by inaccuracies of the estimated model and the relatively small change between provided and corrected GT points randomly changes the error in either direction, either + or -,  by a small amount. 

As expected, the corrected correspondences have zero cross-validation (X-val) error -- all the corrected points are consistent with an $\textbf{H}$ or $\textbf{F}$ model, and this model is recovered in this pseudo-noise free setting, regardless of the point left out.
For $\textbf{H}$ estimation, the errors $\varepsilon_{\text{avg}}, \varepsilon_{\text{med}}$ dropped by about 0.1-0.2 pixels, which is a reasonable value for the positional noise of GT points. For PhotoTour, the GT points were selected from image correspondences perfectly fitting a model estimated from hundreds of points; their correction is minimal. For Kusvod2, the error is reduced by 0.01-0.07 pixels. Note that this is a 1D geometric error w.r.t. $\textbf{F}$, not euclidean in 2D as in homography estimation. These results confirm that the cross-validation error, X-val (provided) is a loose upper bound on the real error.

The ordering of the methods used for homography estimation became clearer than one the provided ground truth points -- VSAC with MAGSAC++ (VSAC$_{\text{MGSC}}$) is always the most accurate and MAGSAC++ is then second most accurate method.
For fundamental matrix estimation, ORSA provides the most accurate results on the PhotoTour dataset, but the difference is negligible, only 0.01-0.02 of a pixel w.r.t. VSAC$_{\text{MGSC}}$ which is the second most accurate algorithm.
On Kusvod2, VSAC$_{\text{MGSC}}$ has the lowest errors. 

\section*{Gauss Elimination for Fundamental Matrix}

The estimation of the fundamental matrix from seven point correspondences, consists of two main steps.
First, constraint $p_2^\text{T} \textbf{F} p_1 = 0$ that each correspondence imply is used to build a linear system $A f = 0$, where $p_i$ is the point in the $i$th image, $\textbf{F}$ is the fundamental matrix, $A$ is the coefficient matrix of the system and $f$ contains the elements of $\textbf{F}$ in vector form~\cite{Hartley2004}.
Coefficient matrix $A$ is of size $7 \times 9$. 
Gaussian Elimination is then used to make $A$ an upper triangular matrix as follows: \\
\[
\begin{pmatrix}
a_{11} & a_{12} & a_{13} & a_{14} & a_{15} & a_{16} & a_{17} & a_{18} & a_{19} \\
0 & a_{22} & a_{23} & a_{24} & a_{25} & a_{26} & a_{27} & a_{28} & a_{29} \\
0 & 0 & a_{33} & a_{34} & a_{35} & a_{36} & a_{37} & a_{38} & a_{39} \\
0 & 0 & 0 & a_{44} & a_{45} & a_{46} & a_{47} & a_{48} & a_{49} \\
0 & 0 & 0 & 0 & a_{55} & a_{56} & a_{57} & a_{58} & a_{59} \\
0 & 0 & 0 & 0 & 0 & a_{66} & a_{67} & a_{68} & a_{69} \\
0 & 0 & 0 & 0 & 0 & 0 & a_{77} & a_{78} & a_{79} \\
\end{pmatrix}.
\] \\
Since the fundamental matrix has 8 degrees-of-freedom the two null-vectors can have the last element fixed to one as $f^{(1)}_9  = f^{(2)}_9 = 1$. 

Let us for the first null-vector fix the eighth element to zero $f^{(1)}_8 = 0$, thus, seventh element becomes $f^{(1)}_7 = -a_{79} / a_{77}$. 
Similarly, for the second null-vector the seventh element can be fixed to zero $f^{(2)}_7 = 0$ and, thus, the eighth one is $f^{(2)}_8 = -a_{79} / a_{78}$.

All other values of null-vectors can be found by substituting the previously found elements:
\begin{equation}
    f^{(\{1,2\})}_i = \frac{-1}{a_{ii}} \sum_{j = i+1}^{9} f^{(\{1,2\})}_j a_{ij} \:\:\:\:\:\:\:\:\:\: \forall i \in \{1, \dots, 6\}
\end{equation}\\
%
The final fundamental matrix is $f = \alpha f^{(1)} + (1-\alpha) f^{(2)}$. 
\begin{table*}[hbt]
\renewcommand{\arraystretch}{1.12}
\setlength{\tabcolsep}{3.0pt}
\hspace{0.75cm}
\begin{tabular}{|c|c|l|l|r|r|r|}
\hline\rowcolor{headergray}
& & \multicolumn{1}{c|}{Method}& \multicolumn{1}{c|}{GT} & \multicolumn{1}{c|}{$\varepsilon_{\text{med}}$} & \multicolumn{1}{c|}{$\varepsilon_{\text{avg}}$} & \multicolumn{1}{c|}{$\varepsilon_{\text{max}}$} \\ \hline
\multirow{36}{*}{\rotatebox{90}{\textbf{Homography}}} & \multirow{18}{*}{\rotatebox{90}{HPatches \hspace{2ex}(142 pairs)}}
& \multirow{2}{*}{VSAC} & Provided & 0.66 & 0.99 & 5.83 \\ \cline{4-7} 
& & & \cclr Corrected &\cclr0.45 & \cclr0.82 & \cclr6.04 \\ \cline{3-7}
& & \multirow{2}{*}{VSAC$_{\text{MGSC}}$} & Provided & 0.65 & 0.82 & 3.78 \\ \cline{4-7} 
& & & \cclr Corrected & \best 0.41 & \best 0.62 & \best 3.56 \\ \cline{3-7}
& & \multirow{2}{*}{USACv20} & Provided & 0.66 & 0.92 & 4.05\\ \cline{4-7} 
& & & \cclr Corrected & \cclr0.47 & \cclr0.73 &  \snd 4.16 \\ \cline{3-7}
& & \multirow{2}{*}{USAC} & Provided & 0.67 & 5.11 & 370.28\\ \cline{4-7}
& & & \cclr Corrected & \cclr0.56 &\cclr 5.00 & \cclr384.48\\ \cline{3-7}
& & \multirow{2}{*}{OpenCV} & Provided & 0.76 & 1.25 & 10.10 \\ \cline{4-7} 
& & & \cclr Corrected & \cclr0.62 & \cclr1.09 & \cclr9.94 \\ \cline{3-7}
& & \multirow{2}{*}{GC} & Provided & 0.74 & 1.12 & 11.42\\ \cline{4-7} 
& & & \cclr Corrected &\cclr 0.52 & \cclr0.89 & \cclr11.28 \\ \cline{3-7}
& & \multirow{2}{*}{MGSC++} & Provided & 0.66 & 0.86 & 4.91 \\ \cline{4-7} 
& & & \cclr Corrected & \snd 0.42 & \snd 0.64 & \cclr4.81\\ \cline{3-7}
& & \multirow{2}{*}{ORSA} & Provided & 0.75 & 55.74 & 1105.82\\ \cline{4-7} 
& & & \cclr Corrected & \cclr0.76 & \cclr54.42 & \cclr1104.78\\ \cline{3-7}
& & \multicolumn{4}{l}{} & \\ 
& & \multicolumn{4}{l}{} &\\ \cline{3-7}
& & \multirow{2}{*}{\textbf{X-val}} & Provided & 0.58 & 0.71 & 6.94 \\ \cline{4-7} 
& & & \cclr Corrected & \cclr0.00 & \cclr0.00 & \cclr0.00 \\ \cline{2-7} 
& & & &  &  &  \\[-1em] \cline{2-7}

 & \multirow{18}{*}{\rotatebox{90}{EVD \hspace{2ex}(10 pairs)}}
& \multirow{2}{*}{VSAC} & Provided & 3.23 & 3.62 & 8.99 \\ \cline{4-7} 
& & & \cclr Corrected &\cclr3.07 & \cclr3.51 & \cclr9.92 \\ \cline{3-7}
& & \multirow{2}{*}{VSAC$_{\text{MGSC}}$} & Provided & 2.80 & 3.37 & 7.05 \\ \cline{4-7} 
& & & \cclr Corrected & \best 2.51 & \best 3.27 & \snd 9.25 \\ \cline{3-7}
& & \multirow{2}{*}{USACv20} & Provided & 3.26 & 3.78 & 10.88 \\ \cline{4-7} 
& & & \cclr Corrected & \cclr3.00 & \cclr3.53 & \cclr11.76 \\ \cline{3-7}
& & \multirow{2}{*}{USAC} & Provided & 6.56 & 117.73 & 474.08\\ \cline{4-7} 
& & & \cclr Corrected & \cclr6.31 & \cclr130.14 & \cclr485.75\\ \cline{3-7}
& & \multirow{2}{*}{OpenCV} & Provided & 3.68 & 4.53 & 8.80 \\ \cline{4-7} 
& & & \cclr Corrected & \cclr3.55 & \cclr4.22 & \best 9.16 \\ \cline{3-7}
& & \multirow{2}{*}{GC} & Provided & 3.72 & 4.17 & 13.28 \\ \cline{4-7} 
& & & \cclr Corrected & \cclr3.49 & \cclr4.18 & \cclr16.84 \\ \cline{3-7}
& & \multirow{2}{*}{MGSC++} & Provided & 2.85 & 3.51 & 7.99  \\ \cline{4-7} 
& & & \cclr Corrected & \snd 2.56 & \snd 3.41 & \cclr10.66\\ \cline{3-7}
& & \multirow{2}{*}{ORSA} & Provided & 143.69 & 170.65 & 438.44\\ \cline{4-7} 
& & & \cclr Corrected & \cclr190.48 & \cclr181.46 & \cclr482.97\\ \cline{3-7}
& & \multicolumn{4}{l}{} & \\  
& & \multicolumn{4}{l}{} & \\ \cline{3-7}
& & \multirow{2}{*}{\textbf{X-val}} & Provided & 1.79 & 1.80 & 2.29 \\ \cline{4-7} 
& & & \cclr Corrected & \cclr0.00 &\cclr 0.00 & \cclr0.00\\ \hline
\end{tabular}\;\;%
\begin{tabular}{|c|c|l|l|r|r|r|}
\hline\rowcolor{headergray}
& & \multicolumn{1}{c|}{Method} & \multicolumn{1}{c|}{GT} & \multicolumn{1}{c|}{$\varepsilon_{\text{med}}$} & \multicolumn{1}{c|}{$\varepsilon_{\text{avg}}$} & \multicolumn{1}{c|}{$\varepsilon_{\text{max}}$} \\ \hline
\multirow{36}{*}{\rotatebox{90}{\textbf{Fundamental matrix}}}
& \multirow{18}{*}{\rotatebox{90}{PhotoTour \hspace{.7ex}(500 pairs)}}
& \multirow{2}{*}{VSAC} & Provided & 0.16 & 0.18 & 0.80  \\ \cline{4-7} 
& & & \cclr Corrected & \cclr 0.16 & \cclr0.18 &\cclr 0.82 \\ \cline{3-7}
& & \multirow{2}{*}{VSAC$_{\text{MGSC}}$} & Provided & 0.15 & 0.17 & 0.75 \\ \cline{4-7} 
& & & \cclr Corrected &\snd 0.15 & \snd 0.17 & \snd 0.73\\ \cline{3-7}
& & \multirow{2}{*}{USACv20} & Provided &0.17 & 0.22 & 3.44\\ \cline{4-7} 
& & & \cclr Corrected & \cclr0.17 & \cclr0.21 & \cclr3.43 \\ \cline{3-7}
& & \multirow{2}{*}{USAC} & Provided & 0.42 & 0.63 & 8.01 \\ \cline{4-7} 
& & & \cclr Corrected & \cclr0.42 & \cclr0.63 & \cclr8.03\\ \cline{3-7}
& & \multirow{2}{*}{OpenCV} & Provided & 0.39 & 0.73 & 25.25\\ \cline{4-7} 
& & & \cclr Corrected & \cclr0.39 & \cclr0.73 & \cclr25.24\\ \cline{3-7}
& & \multirow{2}{*}{GC} & Provided &0.16 & 0.25 & 13.31\\ \cline{4-7} 
& & & \cclr Corrected & \cclr0.16 & \cclr0.25 & \cclr13.31\\ \cline{3-7}
& & \multirow{2}{*}{MGSC++} & Provided & 0.20 & 0.23 & 1.49\\ \cline{4-7} 
& & & \cclr Corrected & \cclr0.20 & \cclr0.23 & \cclr1.48\\ \cline{3-7}
& & \multirow{2}{*}{ORSA} & Provided & 0.14 & 0.15 & 0.64\\ \cline{4-7} 
& & & \cclr Corrected & \best 0.14 & \best 0.15 & \best 0.63\\ \cline{3-7}
& & \multirow{2}{*}{NG-RSC} & Provided & 0.17 & 0.18 & 1.60\\ \cline{4-7} 
& & & \cclr Corrected &   \cclr0.17 & \cclr 0.18 & \cclr 1.60\\ \cline{3-7}
& & \multirow{2}{*}{\textbf{X-val}} & Provided & 0.06 & 0.06 & 0.16 \\
& & & \cclr Corrected & \cclr0.00 & \cclr0.00 & \cclr0.00\\ \cline{2-7}
& & & &  &  &  \\[-1em] \cline{2-7} 

& \multirow{18}{*}{\rotatebox{90}{Kusvod2 \hspace{2ex}(15 pairs)}}
& \multirow{2}{*}{VSAC} & Provided & 0.55 & 0.77 & 3.47  \\ \cline{4-7} 
& & & \cclr Corrected &\cclr0.51 & \snd0.74 & \best 3.47 \\ \cline{3-7}
& & \multirow{2}{*}{VSAC$_{\text{MGSC}}$} & Provided & 0.52 & 0.76 &  3.47 \\ \cline{4-7} 
& & & \cclr Corrected & \best 0.45 & \best 0.73 & \best 3.47 \\ \cline{3-7}
& & \multirow{2}{*}{USACv20} & Provided &0.60 & 1.01 & 5.42\\ \cline{4-7} 
& & & \cclr Corrected &\cclr 0.56 &  0.98 & \snd 5.41 \\ \cline{3-7}
& & \multirow{2}{*}{USAC} & Provided & 2.09 & 2.85 & 15.07\\ \cline{4-7} 
& & & \cclr Corrected & \cclr2.08 & \cclr2.84 & \cclr15.09\\ \cline{3-7}
& & \multirow{2}{*}{OpenCV} & Provided & 1.51 & 6.26 & 63.05 \\ \cline{4-7} 
& & & \cclr Corrected & \cclr1.55 & \cclr6.26 & \cclr63.06 \\ \cline{3-7}
& & \multirow{2}{*}{GC} & Provided & 0.55 & 3.94 & 48.48\\ \cline{4-7} 
& & & \cclr Corrected & \cclr0.54 &\cclr 3.92 &\cclr 48.48\\ \cline{3-7}
& & \multirow{2}{*}{MGSC++} & Provided & 0.58 & 1.18 & 5.69\\ \cline{4-7} 
& & & \cclr Corrected & \cclr0.58 & \cclr1.16 & \cclr5.69\\ \cline{3-7}
& & \multirow{2}{*}{ORSA} & Provided & 0.51 & 14.29 & 307.42\\ \cline{4-7} 
& & & \cclr Corrected & \cclr 0.49 & \cclr14.26 & \cclr307.42\\ \cline{3-7}
& & \multirow{2}{*}{NG-RSC} & Provided & 0.48 & 2.31 & 50.04\\ \cline{4-7} 
& & & \cclr Corrected & \snd 0.46 & \cclr 2.28 & \cclr 50.04 \\ \cline{3-7}
& & \multirow{2}{*}{\textbf{X-val}} & Provided & 0.91 & 1.12 & 2.34 \\ \cline{4-7} 
& & & \cclr Corrected & \cclr0.00 & \cclr0.00 & \cclr0.00 \\ \hline
\end{tabular}
\caption{The median ($\varepsilon_{\text{med}}$), average ($\varepsilon_{\text{avg}}$) and maximum ($\varepsilon_{\text{max}}$) errors in pixels on the used datasets when using the provided ground truth correspondences and the corrected ones projected to the model manifold as reference inliers.
The lowest and second lowest errors are highlighted in red and blue, respectively. }

\label{table:corrected_gt_errors}
\end{table*}

\subsection*{Detecting of pure rotation}
Let $({\bf x}, {\bf x}^{\prime})$ be a point correspondence, 
$\mathbf{K}_1, \mathbf{K}_2$ are the intrinsic camera matrices, 
$\mathbf{R}_1, \mathbf{R}_2$ are the camera rotations, 
$X$ is the unknown 3D object point, and 
the scene has no translation. 
In this case, the following projection equation holds.
\begin{equation}
    {\bf x}\; \sim \mathbf{K}_1 \,\mathbf{R}_1\, X, \quad {\bf x}^{\prime} \sim \mathbf{K}_2\, \mathbf{R}_2\, X,
\end{equation}
where point ${\bf x}^{\prime}$ relates to ${\bf x}$ as follows: 
\begin{equation}
    {\bf x}^{\prime} \sim \mathbf{K}_2\,\mathbf{R}_2\,\mathbf{R}^{\top}_{1}\,\mathbf{K}^{-1}_{1} \,{\bf x},
\end{equation}
where operator $\sim$ means equality up-to-scale.
Homography $\mathbf{H} = \mathbf{K}_{2} \mathbf{R}_2\,\mathbf{R}^{\top}_{1} \mathbf{K}^{-1}_{1}$ transforms image points as
\begin{equation}
    {\bf x}^{\prime} \sim \mathbf{H} \,{\bf x},
\end{equation}
%
In the normalized by $\mathbf{K}_1$ and $\mathbf{K}_2$ points coordinate a homography $\Tilde{\mathbf{H}}$ is conjugated to rotation:
\begin{equation}
     \mathbf{H}^\prime = \mathbf{K}_{2}^{-1} \mathbf{H} \mathbf{K}_1 = \mathbf{R}_2 \mathbf{R}^{\top}_1  = \mathbf{R}.
\end{equation}
\noindent
Once, a homography $\hat{\mathbf{H}}$ with significant support is found, which transforms image correspondences, it is being converted into normalized homography $\hat{\mathbf{H}}^\prime$ via calibration matrices.
If $\hat{\textbf{H}}^{\prime\top} \hat{\textbf{H}}^\prime$ is close (\emph{e.g.,} Frobenius norm) to identity matrix $\mathbf{I}$ then homography is conjugated to rotation matrix, because $\mathbf{R}^\top \mathbf{R} = \mathbf{I}$. Therefore, no translation case is detected.

\end{document}